\documentclass[runningheads]{llncs}
\usepackage{eccv}
\usepackage{eccvabbrv}
\usepackage{booktabs}
\usepackage{fancyhdr}
\usepackage[accsupp]{axessibility}  %
\usepackage{algorithm}
\usepackage{algpseudocode}
\usepackage{booktabs}
\usepackage{bm}
\usepackage[numbers]{natbib}
\usepackage[colorlinks,citecolor=eccvblue]{hyperref}

\usepackage{listings}

\pdfoutput=1

\usepackage{xcolor}
\definecolor{codegreen}{rgb}{0,0.6,0}
\definecolor{codegray}{rgb}{0.5,0.5,0.5}
\definecolor{codepurple}{rgb}{0.58,0,0.82}
\definecolor{backcolour}{rgb}{0.95,0.95,0.92}

\lstdefinestyle{mystyle}{
  basicstyle=\ttfamily\footnotesize,
  breakatwhitespace=false,         
  breaklines=true,                 
  captionpos=b,                    
  keepspaces=true,                 
  numbers=left,                    
  numbersep=5pt,                  
  showspaces=false,                
  showstringspaces=false,
  showtabs=false,                  
  tabsize=2
}
\usepackage{rotating}
\usepackage[rightcaption]{sidecap}
\usepackage{gensymb}
\usepackage{relsize}
\usepackage{makecell}

\usepackage{orcidlink}

\sidecaptionvpos{figure}{c}

\definecolor{turquoise}{cmyk}{0.65,0,0.1,0.3}
\definecolor{purple}{rgb}{0.65,0,0.65}
\definecolor{dark_green}{rgb}{0, 0.5, 0}
\definecolor{orange}{rgb}{0.8, 0.6, 0.2}
\definecolor{red}{rgb}{0.8, 0.2, 0.2}
\definecolor{darkred}{rgb}{0.6, 0.1, 0.05}
\definecolor{blueish}{rgb}{0.3, 0.3, .6}
\definecolor{light_gray}{rgb}{0.7, 0.7, .7}
\definecolor{pink}{rgb}{1, 0, 1}
\definecolor{greyblue}{rgb}{0.25, 0.25, 1}
\definecolor{awesome}{rgb}{1.0, 0.13, 0.32}

\definecolor{figred}{rgb}{0.9, 0.1, 0.1}
\definecolor{figgreen}{rgb}{0.1, 0.7, 0.1}
\definecolor{figblue}{rgb}{0.1, 0.1, 0.9}
\definecolor{figmagenta}{rgb}{0.8, 0.1, 0.8}

\begin{document}

\title{EraseDraw: Learning to Insert Objects by Erasing Them from Images}
\titlerunning{EraseDraw}

\author{Alper Canberk \and
Maksym Bondarenko \and
Ege Ozguroglu \and
Ruoshi Liu \and
Carl Vondrick
}
\institute{Columbia University \\ 
{\larger \href{https://erasedraw.cs.columbia.edu/}{erasedraw.cs.columbia.edu}}}

\authorrunning{A. Canberk et al.}

\maketitle

\begin{abstract}
Creative processes such as painting often involve creating different components of an image one by one. Can we build a computational model to perform this task? Prior works often fail by making global changes to the image, inserting objects in unrealistic spatial locations, and generating inaccurate lighting details. We observe that while state-of-the-art models perform poorly on object insertion, they can remove objects and erase the background in natural images very well. Inverting the direction of object removal, we obtain high-quality data for learning to insert objects that are spatially, physically, and optically consistent with the surroundings. With this scalable automatic data generation pipeline, we can create a dataset for learning object insertion, which is used to train our proposed text-conditioned diffusion model. Qualitative and quantitative experiments have shown that our model achieves state-of-the-art results in object insertion, particularly for in-the-wild images. We show compelling results on diverse insertion prompts and images across various domains. In addition, we automate iterative insertion by combining our insertion model with beam search guided by CLIP.

\end{abstract}

\section{Introduction}\label{sec:intro}

Inserting objects into an image based on a language prompt is a challenging task but has many applications in image editing and content creation in general. There are many fundamental challenges in solving this task. Inserted objects need to appear at physically plausible spatial locations that respect the natural distribution of images. Existing objects in the scene must be preserved. The appearance of the inserted objects needs to be consistent with that of the context. The lighting details should match the environmental lighting. Any one of these is traditionally challenging in computer vision, let alone combined.

\begin{figure}[htbp]
  \centering
  \includegraphics[width=\linewidth]{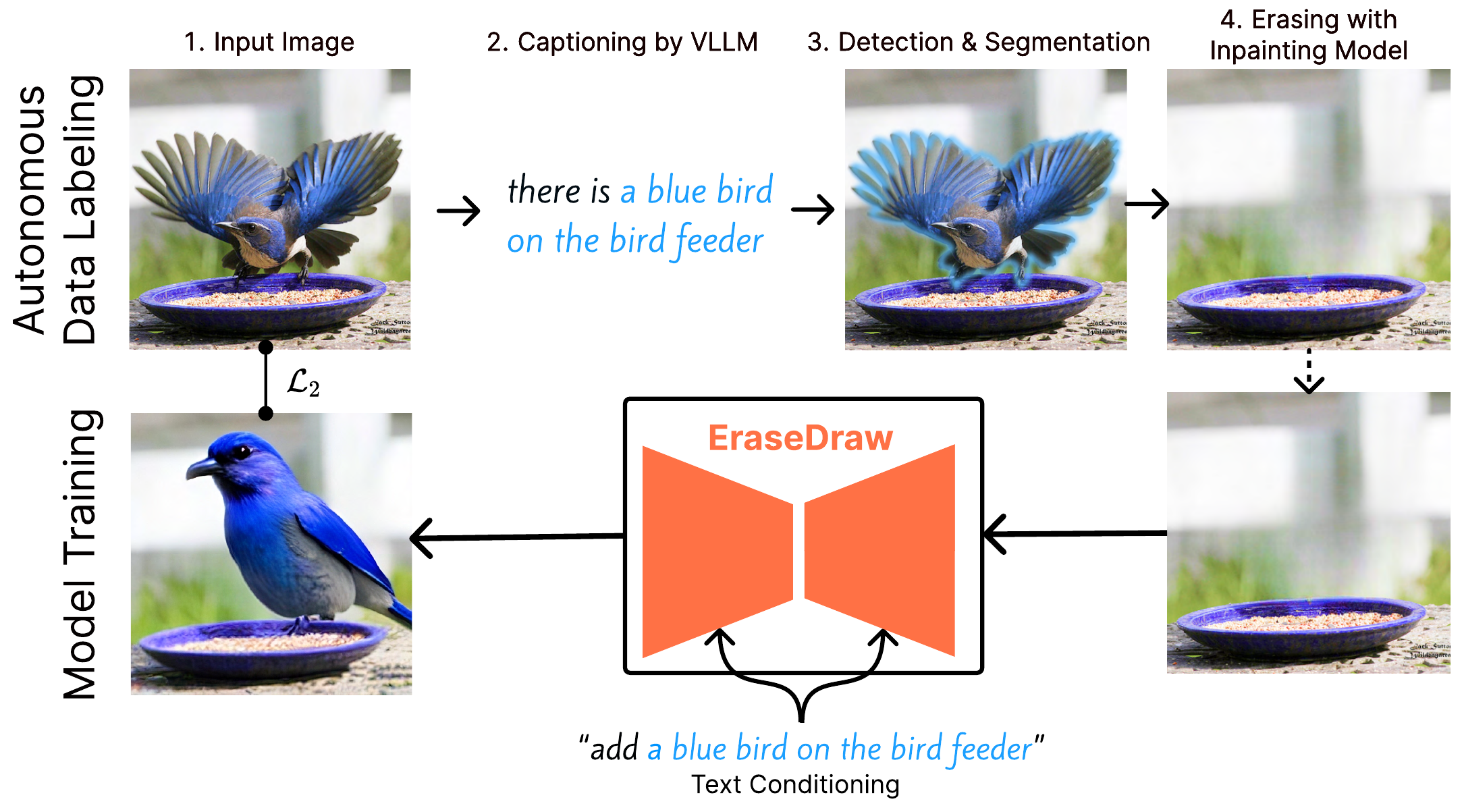}
  \caption{
  \textbf{EraseDraw} We leverage advancements in image understanding and inpainting to train a model that can insert an object given a language instruction. 
  }
  \label{fig:teaser}
\end{figure}

Prior works represented by InstructPix2Pix~\cite{brooks2023instructpix2pix} approached this task by generating pairs of images from pairs of prompts using Prompt2Prompt~\cite{hertz2022prompt}, which is subsequently used to train a model for language-conditioned image editing tasks. While achieving impressive results on image editing tasks such as changing styles or settings, these methods often fail on the tasks of object insertion by making global edits to the image, replacing an existing object when inserting new ones, and struggling to spatially reason (see Figure \ref{fig:failures}). Another line of work attempts to create the training data by rendering from a simulation environment and synthetic objects~\cite{michel2024object}. However, a significant sim2real gap exists, preventing it from generalizing well to in-the-wild images.

In this paper, we propose EraseDraw, a scalable system for learning the task of language-conditioned object insertion into images. We observe that with modern segmentation, captioning and inpainting models, we can perform the task of object removal with much higher photo and physical realism than object insertion. With this observation, we propose an autonomous data generation pipeline for generating input-output pairs for the task of instruction-guided object insertion. We modify a wide distribution of images from the internet by erasing objects from them using inpainting models and describing the attributes and locations of the erased objects using vision language models. As a result, we created a large-scale dataset of paired images as well as language prompts describing the object insertion.

With this dataset, we train a language-conditioned diffusion model to perform the task of object insertion. Experiments show that our model achieves state-of-the-art results, outperforming baselines trained with orders of magnitude more computing resources. Due to our autonomous data generation pipeline, we are able to create training data directly from in-the-wild images, resulting in high-quality object insertion data.

The primary contribution of this paper is a system for learning to insert objects in images and a scalable pipeline for generating the training data for this task from natural images. Insertion furthermore equips us with the ability to plan the iterative generation of novel images in a ``step-by-step'' manner. We believe the ability to insert objects into natural images will have a significant impact on content creation, as well as other related areas, including computer graphics, AR/VR, and robotics.

\section{Related Work}
\label{sec:rel}
\subsection{Image Editing} \label{related:editing}
Generative models such as GANs \cite{goodfellow2014generative,karras2019style,karras2020analyzing} and diffusion models \cite{ho2020denoising,nichol2021improved,rombach2022high} have enabled various image editing tasks such as style transfer \cite{huang2017arbitrary}, image-to-image translation \cite{isola2017image,zhu2017unpaired}, and latent space manipulation \cite{shen2020interpreting,voynov2020unsupervised}. More recently, text-guided diffusion models \cite{nichol2021glide,ramesh2022hierarchical,rombach2022high,saharia2022photorealistic} have allowed intuitive editing of images based on textual prompts.

Several methods have been proposed to enhance the controllability and precision of text-based image editing, however none of them have data neccesary to perform object insertion. SDEdit \cite{meng2021sdedit} employs a stochastic differential equation for iterative denoising to increase the realism of user-provided pixel edits. Prompt2Prompt \cite{hertz2022prompt2prompt} and Null Text Inversion \cite{mokady2022null} modify cross-attention maps to enable both local and global editing. Imagic \cite{kawar2022imagic}, EDICT \cite{su2022edit} and Plug-and-Play \cite{wang2022pretraining} optimize text embeddings for better alignment between the input image and target description. Text2LIVE \cite{bar2022text2live} and Blended Diffusion \cite{avrahami2022blended} train models to add edit layers or blend edited regions along the diffusion process. Imagen Editor \cite{kawar2022imagen} finetunes the diffusion model by inpainting masked objects.

Image Sculpting \cite{sang2023image} presents a novel framework for editing 2D images by converting objects into 3D, allowing direct manipulation of their geometry, and re-rendering them back into the 2D image.
Emu Edit \cite{avrahami2023emu} introduces a multi-task image editing model that achieves state-of-the-art results in instruction-based editing by training on a wide range of tasks and utilizing learned task embeddings.
MagicBrush \cite{su2023magicbrush} improves instruction-based editing by finetuning InstructPix2Pix on a manually-annotated dataset collected using an online editing tool.

To provide more intuitive editing interfaces, InstructPix2Pix \cite{brooks2023instructpix2pix} introduced an instruction-based editing model trained on a synthetic dataset. MagicBrush \cite{su2023magicbrush} further improved it by finetuning on a manually-annotated dataset. However, these methods still struggle with accurately interpreting and precisely executing editing instructions, especially for object insertions. In contrast, our approach leverages the strength of inpainting models to automatically generate high-quality training data for learning object insertion.

\subsection{Object Insertion}
Determining where to place objects in images is a key problem for many editing tasks. Traditional approaches in computer graphics rely on manual specification \cite{zhang2014scene} or synthetic data-driven methods \cite{fisher2012example}. In computer vision, early work used contextual information to predict likely object locations \cite{choi2012context,lin2013holistic,zhao2011image}.

More recently, deep generative models have been used to learn object placements from data. Compositing GAN \cite{azadi2020compositional} generates realistic object composites by predicting geometric and appearance adjustments. RelaxedPlacement \cite{lee2022relaxed} optimizes for object positions and sizes to satisfy inter-object relationships described in scene graphs. OBJect-3DIT \cite{michel2024object} studies 3D-aware object insertion using language instructions on synthetic data.

However, existing methods are often limited in their ability to handle complex, real-world object placements. Our key insight is that inpainting models can be used to erase objects from real images, providing valuable training data to learn meaningful object placements. By inverting this process, we show that we can train models to realistically insert diverse objects into images based on language instructions.

\subsection{Diffusion Models} ~\label{related:diffusion} Denoising Diffusion Probabilistic Models (DDPM)~\cite{ho2020denoising} have marked their place in computer vision as a preferred generative architecture, thanks to their adeptness in handling multi-modal distributions, ensuring training stability, and offering scalability. First, diffusion models were shown to outperform GANs \cite{goodfellow2014generative} in image generation \cite{dhariwal2021diffusion}, followed by Stable Diffusion (SD) demonstrating the efficient scalability of the approach \cite{rombach2022high}. SD achieved this through training on the internet-scale LAION-5B dataset \cite{schuhmann2022laion}, while employing diffusion in the latent space of a Variational Autoencoder (VAE) \cite{kingma2013auto}. More recently, DDPMs were shown to be effective in video generation too, notably by Stable Video Diffusion \cite{blattmann2023stable} and Sora \cite{videoworldsimulators2024}. 

\begin{figure}[t] 
    \centering
    \includegraphics[width=\textwidth]{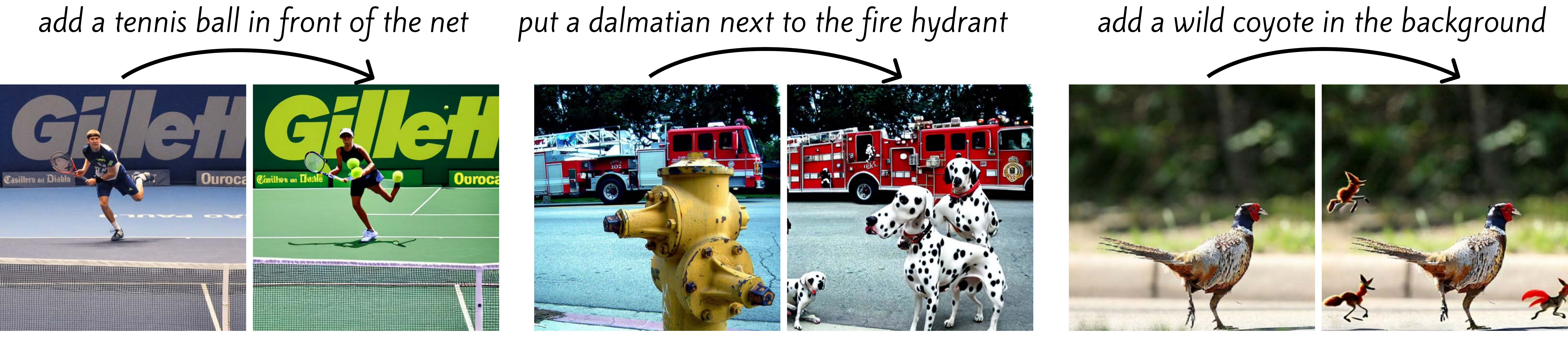}
    \caption{State-of-the-art image editing methods fail to correctly insert objects into visual scenes. They perform global edits that don't preserve scene context (\textbf{left})\cite{brooks2023instructpix2pix}, replace existing objects (\textbf{middle})\cite{su2023magicbrush}, and struggle to spatially reason (\textbf{right}) \cite{hive_magicbrush_checkpoint}. You may see how we did on these examples in Figure \ref{fig:ours-on-failure} of the Appendix.} 
    \vspace{-1em}
    \label{fig:failures}
\end{figure}

With the rise of internet-scale diffusion models, multiple works have studied their excellent ability to represent the natural image manifold, enabling zero-shot generalization in tasks such as 3D-reconstruction \cite{liu2023zero, liu2023syncdreamer, wu2023sin3dm, zeronvs, deitke2024objaverse}, segmentation \cite{xu2023odise, amit2021segdiff}, amodal perception \cite{ozguroglu2024pix2gestalt, zhan2023amodal}, recognition \cite{li2023diffusion, chen2022diffusiondet}, as well as image editing~\cite{brooks2022instructpix2pix,gal2022image,ruiz2023dreambooth}. However, these image-editing methods (\ref{related:editing}) significantly struggle in performing object-insertion. In this work, we leverage the ability of pre-trained diffusion models to represent the natural and multi-modal distribution of people and objects in visual scenes, thereby allowing us to perform object removal first (\ref{sec:method:dataset}), which by inversion enables insertion (\ref{sec:method:diffusion}).

\begin{figure}[t]
    \centering
    \includegraphics[width=\textwidth]{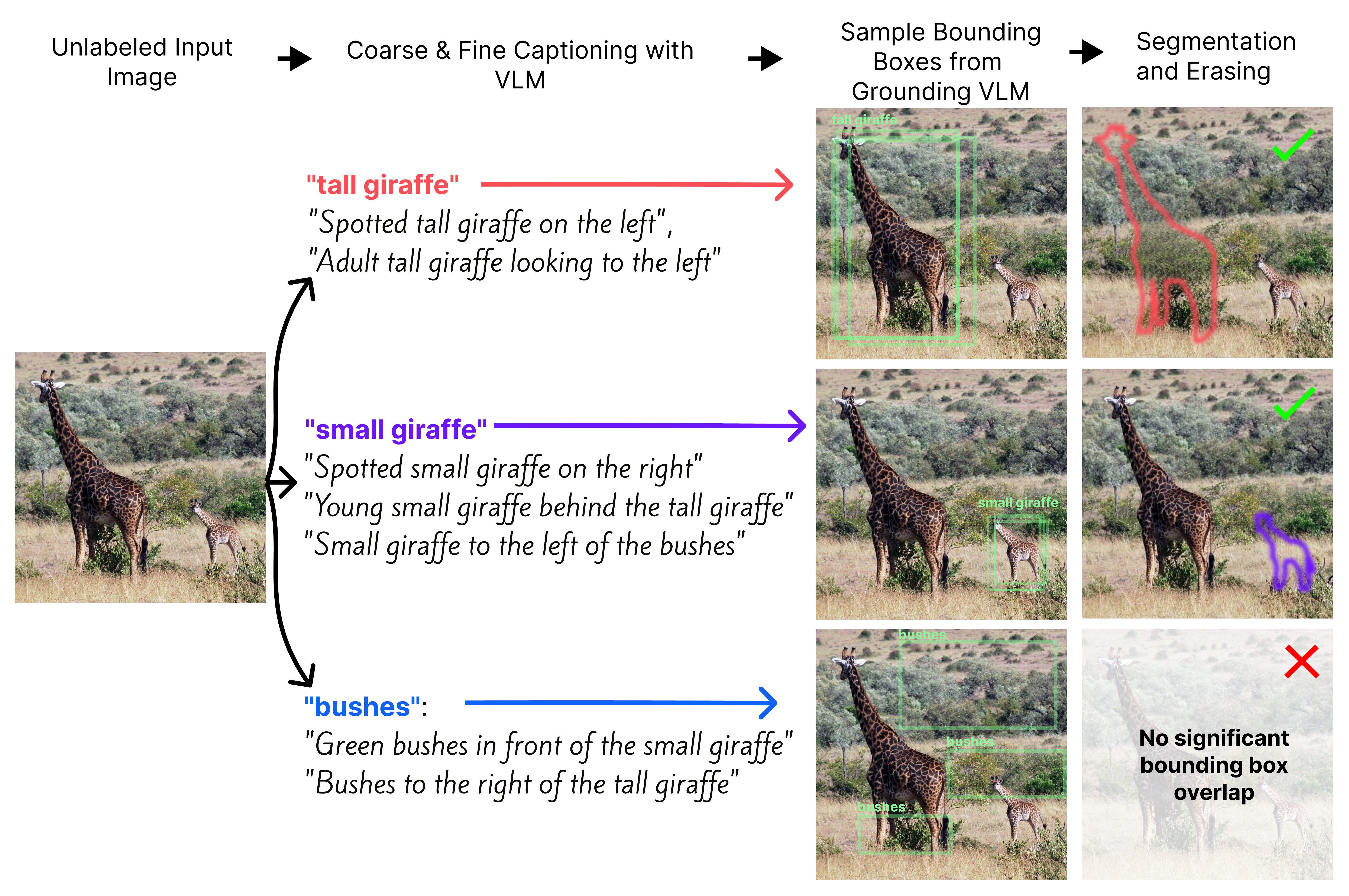} 
    \caption{\textbf{EraseDraw Data Generation Pipeline} (i) An unlabeled image is sampled taken from a dataset (ii) The images are given to a captioning model, which describes the objects in the image (iii) Objects are detected using the coarse caption from the captioning model, and the objects that are confidently detected are (iv) segmented, (v) and erased. The final images are added to the dataset along with the captions corresponding to them. }
    \label{fig:dataset-generation}
\end{figure}

\begin{figure}[t]
    \centering
    \includegraphics[width=\textwidth]{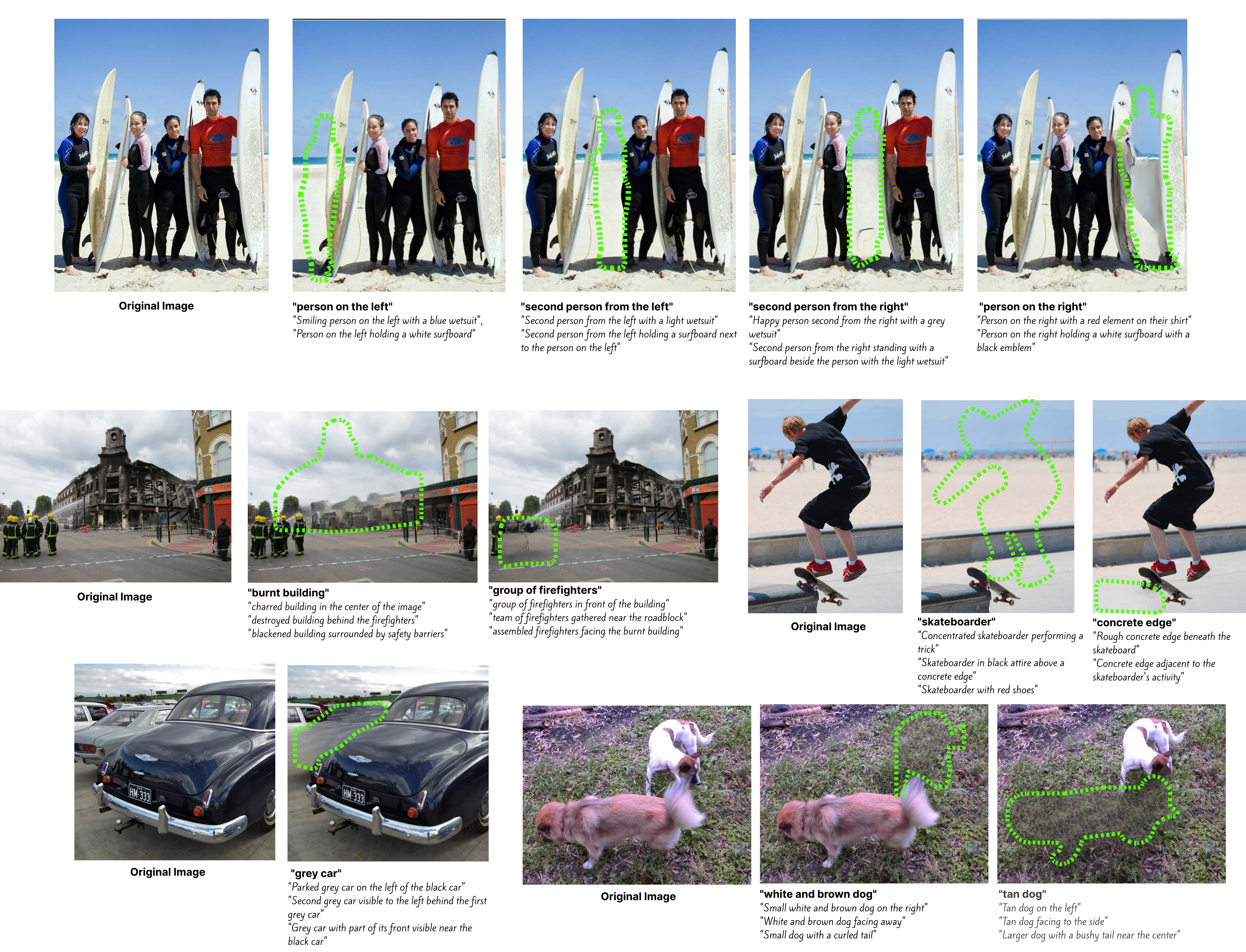} 
    \caption{We show examples from our EraseDraw Dataset.}
    \label{fig:dataset-sample}
\end{figure}

\section{Method}
\label{sec:method}
Our main contribution is a framework for learning object insertion from natural images. This framework brings together the best aspects of existing pre-trained models: combining the powerful descriptive capabilities of multimodal LLMs, complementing it with attribute bound detection abilities of Grounding VLMs, and modifying images with precise segmentation and erasing models. After generating a synthetically annotated dataset of 65,000 images, we fine-tune a large pre-trained diffusion model on our dataset, the procedure for which we outline in Sec.~\ref{sec:method:diffusion}.

\subsection{Automatic Dataset Generation} \label{sec:method:dataset}

For training a language-conditioned object insertion model in a supervised manner, one needs triplets in the form of $(c_I, c_T, \bm{x})$, where $c_I$ represents the source image, $c_T$ describes the identity and location of the object to be inserted, and $\bm{x}$ is modified version of $c_I$, which includes the object described in $c_T$.
Our key insight is that we can let natural images from the internet be the target images $\bm{x}$, and we can autonomously derive the context $c_T$ and $c_I$ by detecting and erasing objects from these images. We illustrate the data generation framework in the following subsections, and we precisely describe our \hyperref[alg:dataset]{algorithm} in the appendix.

\subsubsection{Coarse \& Fine Captioning}

To enhance object identification in images, a powerful Vision-Language Model (VLM), which is GPT-4 \cite{achiam2023gpt} in our implementation, is used to generate both coarse and fine captions for each object. Coarse captions provide a simple, yet unique description focusing on basic attributes like shape and color, aiding in the detection phase. Fine captions offer complex, detailed descriptions, enriching the object's representation. This approach leverages the strength of grounding VLMs, such as CogVLM \cite{wang2023cogvlm}, in detecting objects using straightforward descriptions, while at training time, objects identified through coarse captions can be associated with any of their detailed fine captions. For example, a "red bowl" might be simply identified, but further described in detail as "a dark red bowl on top of the shelf." This methodology ensures efficient object detection while enabling rich, descriptive training data for VLMs.



\subsubsection{Detection with Grounding VLM \& Segmentation} 
Standard open-vocabulary object detectors struggle with accurately identifying colors, sizes, or spatial relationships, leading to inaccuracies in datasets with multiple instances of the same object type. To address this, we utilize CogVLM's  attribute-binding capabilities, which excel in recognizing these features, as demonstrated in Fig.~\ref{fig:dataset-generation} and ~\ref{fig:dataset-sample}. By inputting coarse captions into CogVLM, we generate bounding boxes for objects, though errors can still arise from VLM hallucinating objects or difficulty in identification (e.g., bushes scenario in Fig~\ref{fig:dataset-generation}).

To mitigate these issues, we employ rejection sampling based on CogVLM's uncertainty, using it as a transformer-based probabilistic model. By sampling three bounding boxes at a temperature of 0.2 and accepting those with an Intersection over Union (IoU) exceeding 80\%, we ensure the object's presence and identifiability by the consistency of bounding box locations. Objects with scattered bounding boxes are excluded from the dataset. Identified objects are then segmented using SAM \cite{kirillov2023segment} and prepared for the erasing stage, refining the dataset's accuracy for training purposes.

\subsubsection{Erasing}




We employ the LaMa inpainting model \cite{suvorov2022resolution} for erasing objects using segmentation masks. Although LaMa is less versatile compared to advanced models like Stable Diffusion for inpainting, it is effective in consistently erasing objects. A minor issue with LaMa is artifact creation when erasing large objects, but this does not greatly impact our process due to two main reasons: firstly, images with artifacts serve as conditional inputs rather than direct supervision; secondly, artifacts often diminish after processing through the VAE for latent diffusion, mitigating their presence.



\subsection{Diffusion Model} \label{sec:method:diffusion}

Our goal is to sample from $p(\bm{x}| c_I, c_T)$, the distribution of target images conditioned on a source image and a text instruction. To this end, we train a latent conditional diffusion model $\epsilon_\theta(\cdot)$, which
estimates the score function of $p(\bm{x}| c_I, c_T)$
by optimizing for the simplified variational lower bound objective
\[ \min_\theta \mathbb{E}_{\mathcal{E}(\bm{x}), \mathcal{E}(c_I), c_T, \epsilon \sim \mathcal{N}(0, 1), t}[	\lVert{\bm{\epsilon}_\theta(\bm{z}_t, t, \mathcal{E}(c_I), c_T)} - \bm{\epsilon}\rVert^2_2 ] \]
where $\bm{x}$ is an image sampled from the dataset, $\mathcal{E}(\cdot)$ is a VAE, and $\bm{z}_t = \mathcal{E}(\bm{x})$ is a noisy latent embedding of $\bm{x}$ where the noise level increases with $t \in T$.

In essence, the diffusion model learns to predict the noise $\bm{\epsilon}$ added to a ground-truth latent image $\bm{z}_0$. To sample from the model, we start with pure Gaussian noise tensor and repeatedly invoke the diffusion model to predict and subtract noise, which iteratively denoises the pure noise tensor into a latent image. Finally, we invoke the decoder $\mathcal{D}$ to obtain a full-resolution image from the latent image. 

Following \cite{brooks2023instructpix2pix}, we initialize the weights of our network $\epsilon_\theta$ from Stable Diffusion 1.5, a text-to-image model pretrained on a large-scale dataset of images. Similarly, we allow for conditioning on images by expanding the number of input channels of the convolutional input layer. Initialization from a pretrained model allows our fine-tuned model to generate objects from an open-vocabulary, well beyond the set of objects that are in our fine-tuning dataset of 65,000 images.

\subsection{Iterative Generation and Planning}

Our model's ability to precisely insert an object into an existing image allows the composition of novel images in a "step-by-step" manner.

In particular, given a background image and a sequence of desired edits, one can repeatedly invoke EraseDraw, sequentially prompting the model with each edit, and feeding in the output image of the last step as the input to the next.


\subsubsection{Automating Step-by-Step Composition}\label{automatingiterative}

To compose a complex image by its constituents, a user needs to decide the order in which to insert each object. For each inserted object, the user may also choose to review a multitude of samples from the diffusion model to pick which edited image should be the input to the next step.

Alternatively, the decision-making and verification process can be offloaded to off-the-shelf vision and language models. We instantiate such an application by implementing \textit{beam search} with a CLIP Score heuristic. Starting with an unmodified image and a pre-specified sequence of edits, for each step, we sample $N$ images for each of the $k$ 'beams', resulting in $Nk$ images. Then, we compute text-image CLIP alignment between the $Nk$ generated images and the editing instruction to rank images according to their quality and continue onto the next step with the top $k$.

\begin{figure}[tb] 
  \centering
  \begin{subfigure}[b]{0.49\linewidth} 
    \includegraphics[width=\textwidth]{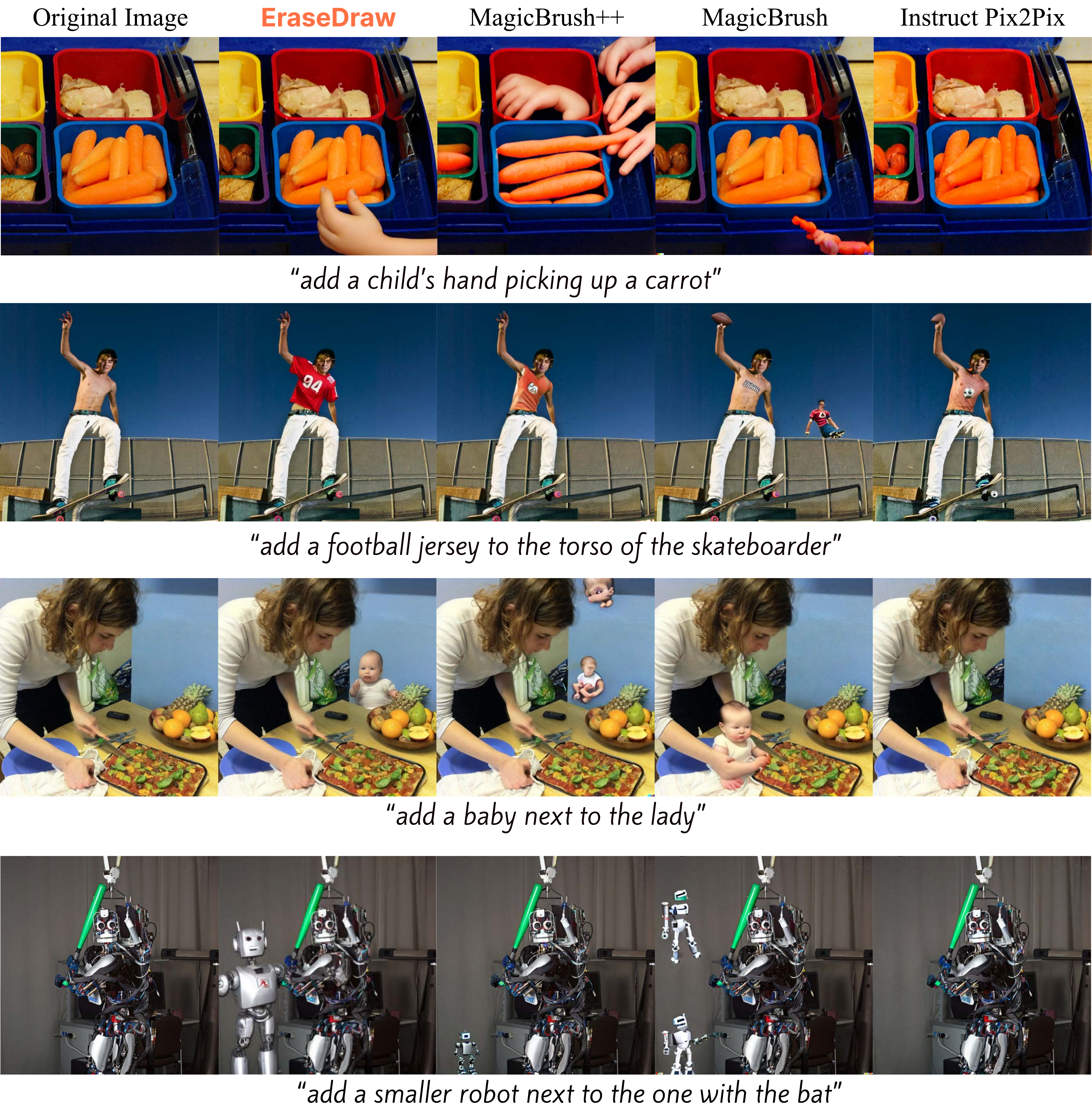}
  \end{subfigure}
  \hfill 
  \begin{subfigure}[b]{0.49\linewidth} 
    \includegraphics[width=\textwidth]{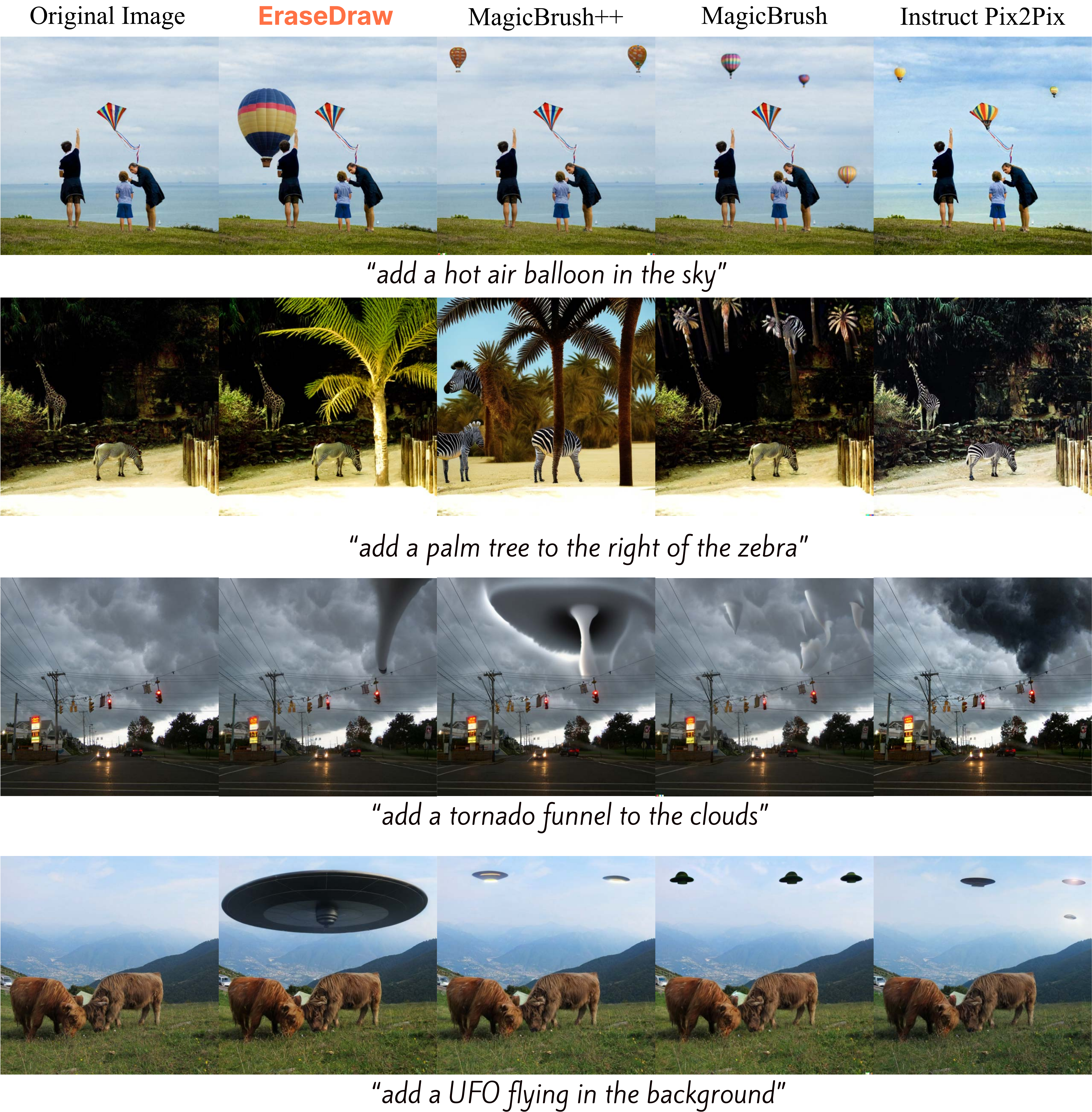}
  \end{subfigure}
  \caption{\textbf{Qualitative Results on EmuEdit Benchmark} on inserting people and outdoor objects.}
  \label{fig:baselines1}
\end{figure}

\begin{figure}[tb]
  \centering
  \begin{subfigure}[b]{0.49\linewidth} 
    \includegraphics[width=\textwidth]{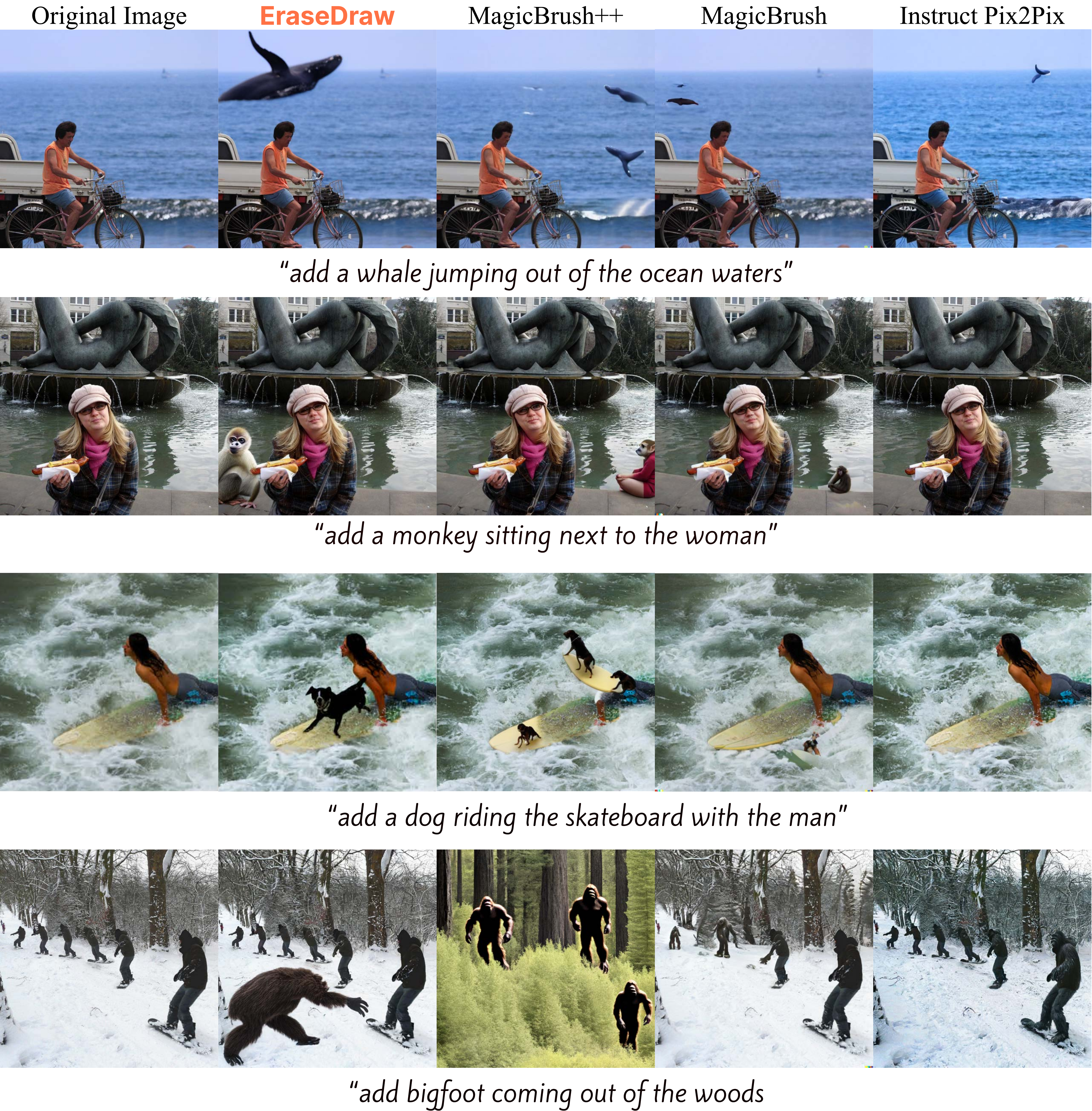}
  \end{subfigure}
  \hfill 
  \begin{subfigure}[b]{0.49\linewidth} 
    \includegraphics[width=\textwidth]{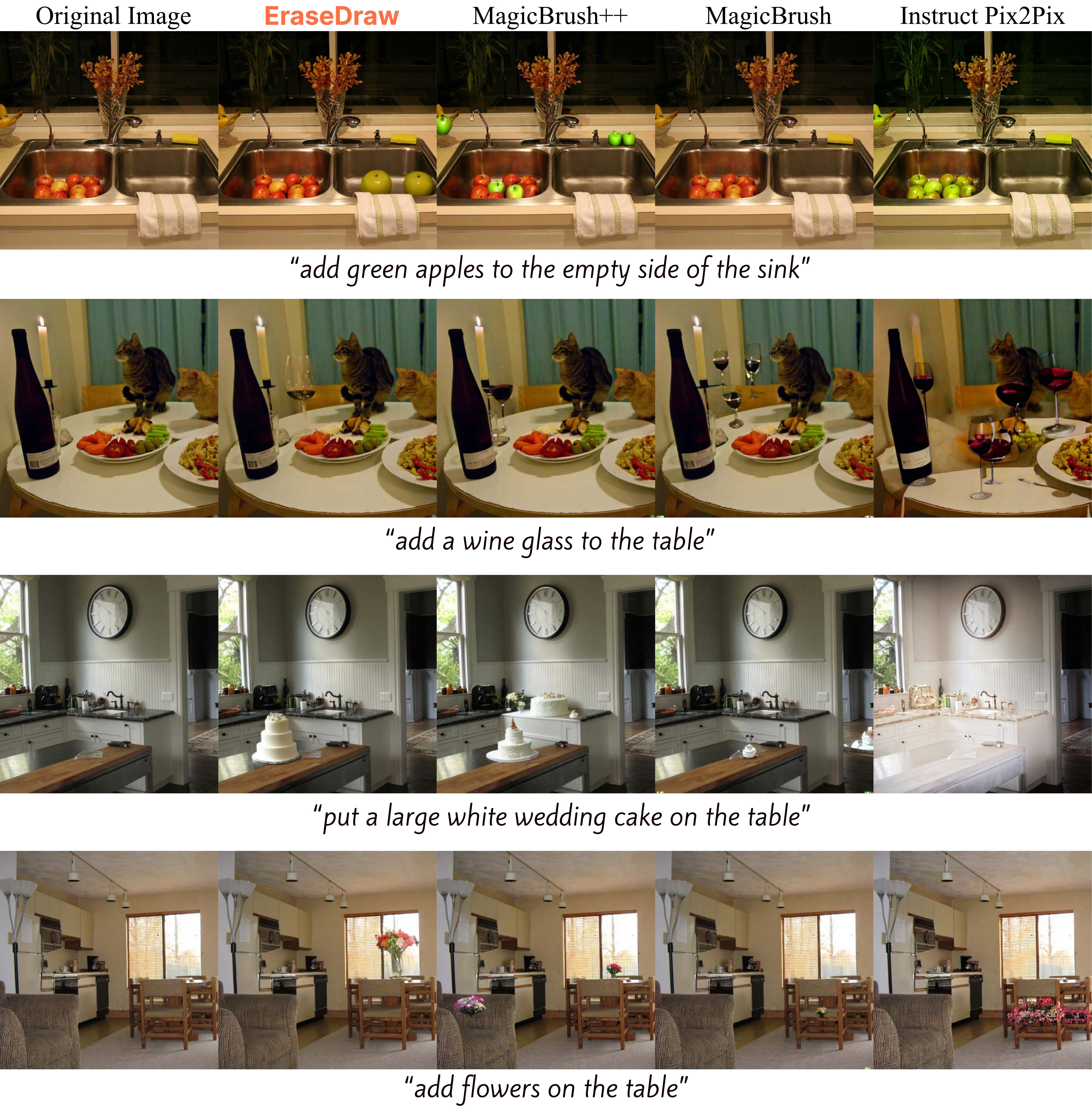}
  \end{subfigure}
  \caption{\textbf{Qualitative Results on EmuEdit Benchmark} on inserting animals and household objects.}
  \label{fig:baselines2}
\end{figure}

The goal of our experiments is to evaluate our model's ability to make precise, high-fidelity insertions that are faithful to the given prompt. To this end, we conduct quantitative and qualitative comparisons with publicly available language-guided image editing models on single-step object insertion (\ref{single_step}). Then, we demonstrate our model's ability to iteratively compose scenes step-by-step (\ref{iterative}). Finally, we present quantitative ablations for our dataset recipe. 

\subsection{Single-Step Object Insertion} \label{single_step}
\subsubsection{Setup} Given an RGB image and a text prompt, the task of single-step object insertion aims to generate a new image with the specified object inserted while faithfully preserving the input scene. We compare the performance of EraseDraw with two similar-architecture models: InstructPix2Pix~\cite{brooks2023instructpix2pix}, trained on 450,000 auto-labeled images, and MagicBrush~\cite{su2023magicbrush}, a fine-tuned version of InstructPix2Pix with ~5,000 human-generated examples. We are not able to compare against EmuEdit~\cite{sheynin2023emu} since the model is not publicly available.


\subsubsection{Metrics} 



For evaluating object insertion, we use a part of EmuEdit's~\cite{sheynin2023emu} validation set designed for "add" instructions, comprising tuples of captions and images before and after edits. Models are judged by $\text{CLIP}_{out}$, comparing output captions with generated images, and $\text{CLIP}_{dir}$, evaluating the alignment between changes in captions and images. Lastly, to confirm our autonomously evaluated results, ten participants performed pairwise ranking tasks across 15 images and 4 models (90 pairwise comparisons per user) on a random subset of the EmuEdit insertion tasks. We report the win rate of each model. 

\begin{table}[t]
\centering

\begin{tabular}{lcccc}
\toprule
Method &   Dataset Size & $\text{CLIP}\textsubscript{out}\uparrow$ &   $\text{CLIP}\textsubscript{dir}\uparrow$ & Win Rate (\%)\\
\midrule
Erasedraw  & 65,000   & \textbf{0.0930} & \textbf{0.1383} & 82.22 \\
Magicbrush++ &  -  &          0.0896 &         0.1304 & 55.06\\
Magicbrush &   455,000  &       0.0854 &         0.0930 &  48.64\\
InstructPix2Pix & 450,000  &          0.0746 &         0.0976 & 12.56\\
\bottomrule
\end{tabular}
\caption{Comparison with language-guided editing baselines on the EmuEdit dataset}
\label{tab:method_comparison}
\end{table}

\subsubsection{Results} 



\begin{figure}[p]
    \centering
    \includegraphics[width=\linewidth]{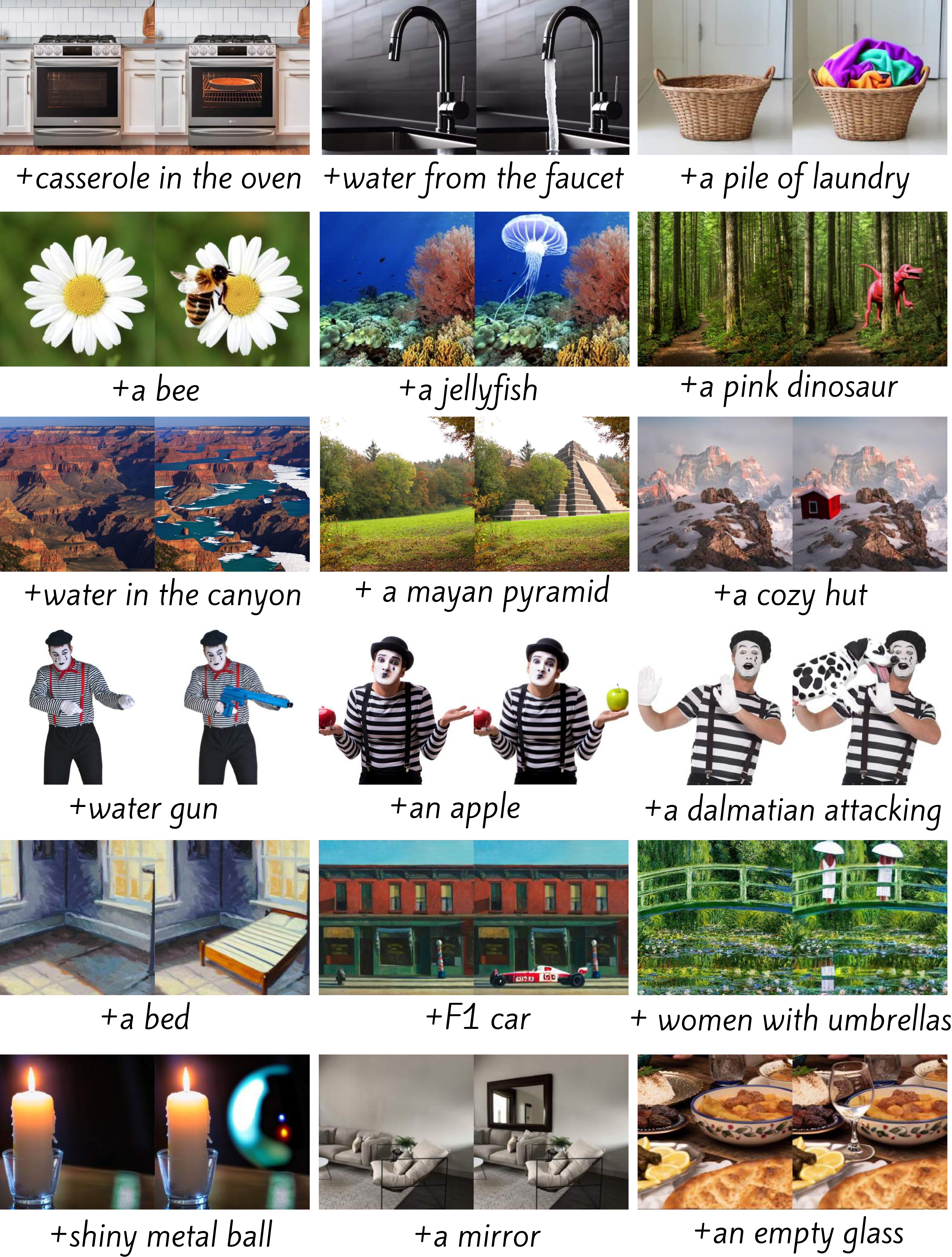} 
    \caption{\textbf{Single-Step Generation.} EraseDraw can perform complex object insertion tasks such as inserting water flowing down a sink, placing objects behind occluders, predicting hand-poses with correct affordance, and accounting for lighting effects.}
    \label{fig:single-step}
\end{figure}

Table $\ref{tab:method_comparison}$ shows the effectiveness of our model on the EmuEdit object insertion benchmark, as indicated by higher $\text{CLIP}_{out}$ and $\text{CLIP}_{dir}$ scores, despite our model being trained on one order of magnitude less data. In Fig.~$\ref{fig:baselines1}$ and $\ref{fig:baselines2}$, we qualitatively contrast our system with baselines, showcasing humans, animals, and indoor and outdoor scenes. These illustrate the nuanced, realistic, and vibrant edits of EraseDraw compared to the baseline models, which face numerous challenges as depicted in Fig.~$\ref{fig:failures}$. In Fig.~\ref{fig:single-step}, we further showcase intricate edits like water flow simulation in a sink, object placement behind existing items, human-object interaction predictions, and lighting adjustments. 

\begin{figure}[t] 
    \centering
    \includegraphics[width=\textwidth]{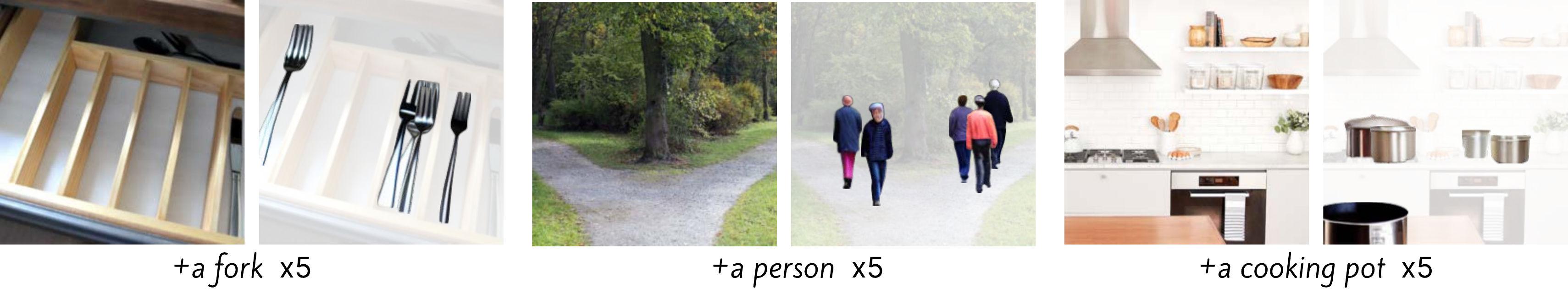}
    \caption{\textbf{Modeling the Multimodal Distribution of People and Objects.} Sampling from our model reveals where objects naturally appear in the world. This opens up applications where commonsense knowledge about object placements are required, such as embodied agents.} 
    \label{fig:naturaldistribution}
\end{figure}

Additionally, in Fig.~$\ref{fig:naturaldistribution}$, we display our model's capability to accurately reflect the natural spatial distribution of objects and people in a given scene. This capability highlights the advantage of learning from natural images, as opposed to simulated objects in prior work ~\cite{michel2024object}.

\subsection{Iterative Generation} \label{iterative}

\subsubsection{Setup} In this section, we display qualitative results of the iterative object insertion process in both human-guided and automated settings. We start with empty background images and iteratively invoke our model with a predetermined sequence of instructions. To compare this process with the conventional one-shot text-to-image method, we combine the set of edit instructions into a single prompt and generate an image from Stable Diffusion 1.5. 

\subsubsection{Results}

\begin{figure}[p]
    \centering
    \includegraphics[width=\linewidth]{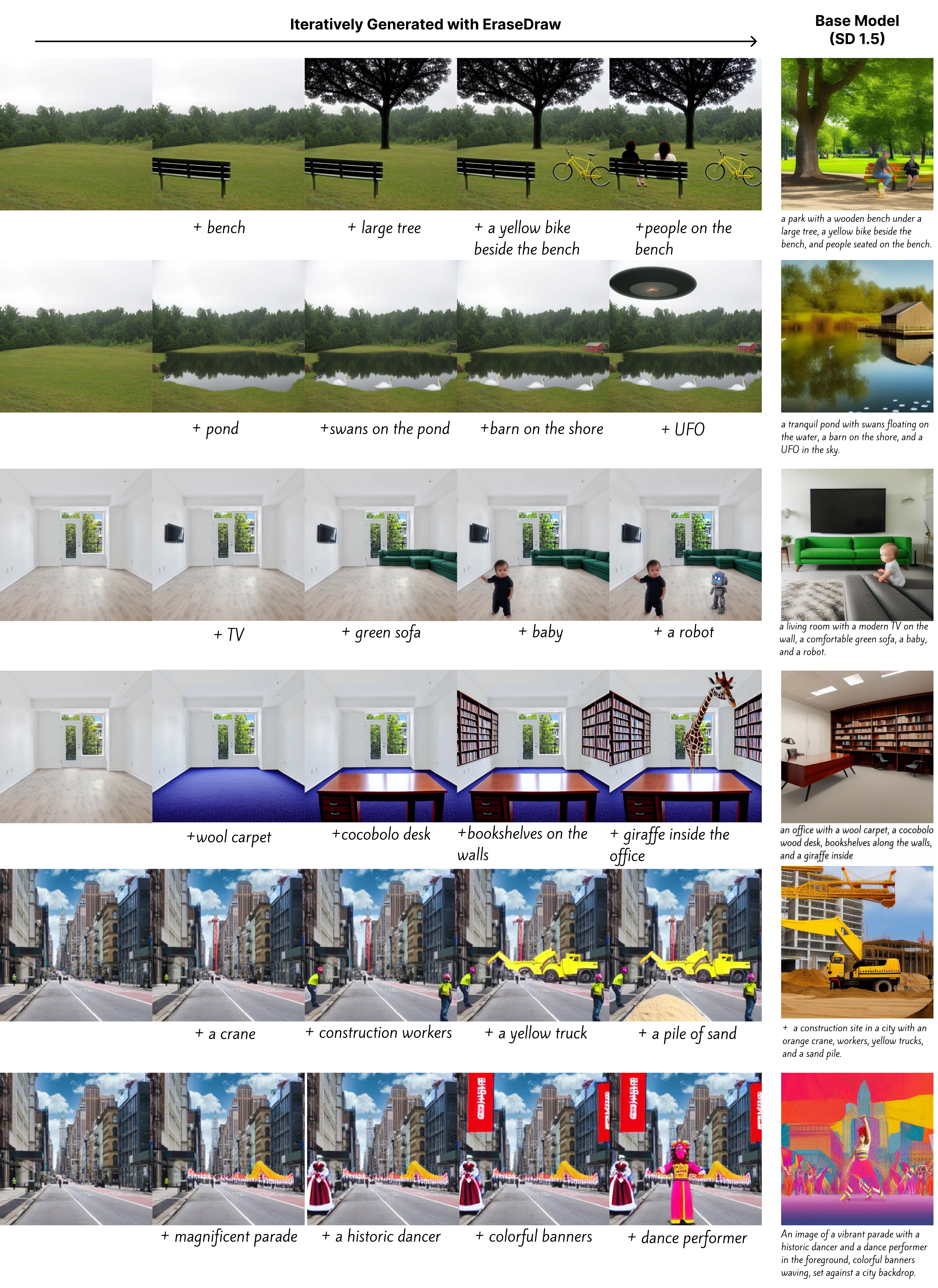} 
    \caption{\textbf{Iterative Generation}}
    \label{fig:iterative-generation}
\end{figure}

According to Fig.~\ref{fig:iterative-generation}, we notice that Stable Diffusion 1.5 fails to follow the entire prompt precisely, confusing attribute bindings or omitting insertions that would appear unnatural in the described context such as the 'giraffe in the office. In comparison, our method results in complex final images that accurately match the edit instructions. This is a surprising result given that our base model is Stable Diffusion 1.5. Iteratively composing the image appears to have addressed important problems of attribute binding and out-of-distribution robustness. We attribute this to the fact that performing a single step of insertion is a much easier task than trying to satisfy all constraints of one-shot text-to-image image generation. 

\begin{figure}[t]
    \centering
    \includegraphics[width=\linewidth]{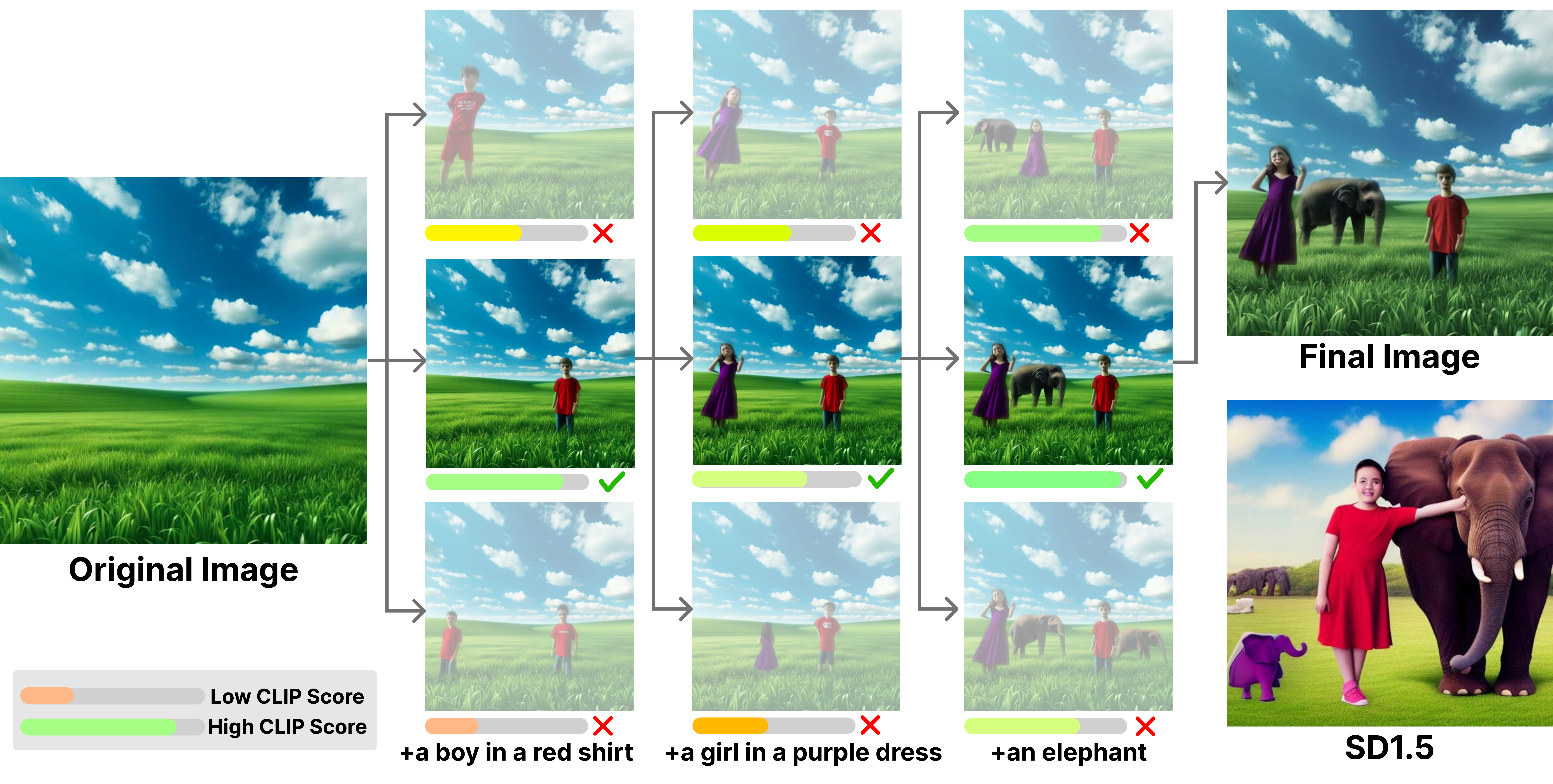} 
    \caption{\textbf{Beam Search} Given the original image on the left, we run beam search using CLIP distance between the edit instruction and the output image as the score heuristic. In this figure, we display the results from the top beam. To compare with one-shot diffusion using SD1.5, we use the prompt "A boy in a red shirt, girl in a purple dress, and an elephant on a grassy field."
    }
    \label{fig:beam-search}
\end{figure}

\subsection{Beam Search}

\subsubsection{Setup} We instantiate the beam search procedure outlined in $\ref{automatingiterative}$ with branching factor $N=4$, beam width of $k=3$, for 3 iterations. For clarity, we only trace and display the top beam and the top-3 generations from that beam at each step.  As a baseline, we sample $TNk$ images from SD1.5 (our base model) using the combined prompt, and present the image with the highest CLIP score.

\subsubsection{Results} We keep the inference setting consistent with Sec.~\ref{iterative}. Fig.\ref{fig:beam-search} shows that CLIP score successfully finds the best candidates for the next step of insertion, resulting in a final image that composes an accurate image without human intervention. Meanwhile, SD1.5 mixes up attributes to their respective objects. While we have shown only 3 steps of edits in these evaluations, the performance between one-shot and iterative image generation increases sharply with additional steps. We include more results in figures \ref{fig:iterative-generation-1}, \ref{fig:iterative-generation-2}, \ref{fig:iterative-generation-3}, \ref{fig:iterative-generation-4}, and \ref{fig:iterative-generation-5}.

\subsection{Dataset Ablation} \label{dataset_ablation}

\subsubsection{Setup}

Finally, we perform ablation studies on our dataset generation pipeline to evaluate its contribution to the results. First, we use our end-to-end data generation pipeline to annotate 15,000 images, which we call 'EraseDraw-15k'. This pipeline could easily be scaled to orders of magnitude more images given enough compute. In addition, we convert 50,000 images from GQA~\cite{hudson2019gqa} with annotated scene graphs into edit instructions by segmentation and erasing. We train EraseDraw on a combination of these two datasets.

\subsubsection{Results}
Despite being autonomously generated and smaller size compared to GQA-Insert, the model trained on EraseDraw-15k competes well with its GQA-Insert counterpart, with the combination of both datasets yielding superior results (see Table \ref{tab:dataset_ablation}). This performance differential can be attributed to the complementary qualities of the datasets: GQA-Insert's human-labeled accuracy versus EraseDraw-15k's broader diversity, albeit with more errors and limited volume.


\begin{table}[b] \label{tab:dataset_ablation}
\centering
\begin{tabular}{lcccc}
\toprule
Dataset & \multicolumn{1}{c}{\begin{tabular}[c]{@{}c@{}}Dataset\\ Size\end{tabular}} & \multicolumn{1}{c}{\begin{tabular}[c]{@{}c@{}}Autonomously\\ Generated\end{tabular}} & \text{CLIP}\textsubscript{out}\(\uparrow\) & \text{CLIP}\textsubscript{dir}\(\uparrow\) \\
\midrule
GQA-Insert & 50,000 & No & 0.0907 & 0.1279 \\
EraseDraw-15k & 15,000 & Yes & 0.0905 & 0.1254  \\
Mixed & 65,000 & No & \textbf{0.0930} & \textbf{0.1383}  \\
\bottomrule
\end{tabular}
\caption{EraseDraw Dataset Ablation Studies}
\label{tab:dataset_ablation}
\end{table}

\section{Discussion} \label{sec:conclusion}

\subsection{Conclusion and Limitations}
We introduce EraseDraw, a framework for autonomously generating object insertion data by leveraging the fact that erasing is an easier task than drawing. We train a diffusion model using our data, and we show that it achieves state-of-the-art object insertion results. Additionally, we introduce a new paradigm for iteratively composing an image using object insertion steps. We emphasize that our approach is model-agnostic; all general-purpose image editing models can benefit from including object insertion examples from EraseDraw in their training data. Besides image generation and editing, EraseDraw can unlock downstream applications such as giving robots a commonsense understanding of object placements and augmenting existing image datasets with novel objects.

Our work is limited by the relatively small created dataset and the base model. In particular, we note that the quality of the inserted objects is limited by the capabilities of the base model we finetune from, consequently our model is unable to generate anatomically accurate hands or faces, and it has limited ability to follow instructions involving precise object placements. Furthermore, while we our framework can be used for processing arbitrary unlabeled images, our dataset is derived from OpenImages and COCO, which may have biases for certain object categories.

We believe scaling up our autonomous data pipeline and fine-tuning stronger base models will yield superior performance.

\subsection{Societal Impact}
Image editing tools such EraseDraw could potentially be misused for creating misleading or harmful content. It is crucial for the research community to develop techniques for mitigating misuse, and to promote responsible use of these technologies.

\vspace{3mm}
{\small
\textbf{Acknowledgements:}
The authors would like to thank for Huy Ha, Zeyi Liu, Samir Gadre and the Columbia CV Lab valuable feedback and discussions.
}

\renewcommand{\bibsection}{\section*{References}}


\renewcommand\thesection{\Alph{section}}
\renewcommand\thesubsection{\thesection.\arabic{subsection}}

\newpage
{
\centering
\Large
\textbf{EraseDraw: Learning to Insert Objects by Erasing Them from Images} \\
\vspace{0.5em}Supplementary Material \\
\vspace{1.0em}
}
\appendix

\section{Training}
We train our model for 1000 steps on 2 $\times$ 48GB A6000 GPUs over 12 hours at $256 \times 256$ resolution. With the help of gradient accumulation, we train at an effective batch size of 1536 images. We adopt all other model related training hyperparameters from \cite{brooks2023instructpix2pix}. We find that our model can generalize to non-square images and resolutions as high as $512 \times 512$

\section{Accelerating Inference}

While the inference duration of diffusion models is prohibitive, we accelerate our model using a pre-trained LCM-LoRA \cite{luo2023lcm} for Stable Diffusion 1.5. Surprisingly, despite being trained for the text-to-image task, the LCM-LoRA preserves the insertion abilities of our model while cutting down the number of inference steps to as low as 4 to 16 steps depending on the desired image quality. 

All results in our paper except our quantitative evaluations and qualitative comparison figures have been generated using LCM LoRA with 16 steps using 512 $\times$ 512 resolution. 

Our qualitative comparison figures are generated with 50 steps of Euler ancestral sampling outlined in Karras et al. \cite{karras2022elucidating} with 512 $\times$ 512 resolution.

\section{Prompts}

\subsubsection{Extracting Captions from Images using Large Vision Language Model}

To extract captions from an image using GPT-4 Vision, we use the following prompt

\begin{lstlisting}
1. Describe the objects in this image along with their attributes and their spatial relations with respect to the other objects.
2. For every individual object,
    a) Come up with a "subject identification" for that object. The subject identification should be a simple way to identify the object in the image. Color and shape may be helpful to include.
    Examples:
        If two fish are in the picture, one is red and the other is blue, then you could identify their subject identificaitons as "the red fish" and "the blue fish".
        If three men are in the picture, and they're spread out into a row, then you could identify their subject identifications as "man on the left","man on the right",man in the middle".
        If only a single cat is in the picture, then the subject identification "cat" is sufficient.
    The subject identification MUST NOT include a noun other than the subject. 
    b) Come up with simple captions of the form [adjective] [subject] [prepositional phrase] that describe the location of the object with respect to other objects or the image.
    (e.g. a man with a blue shirt standing in front of the wall, an elephant next to the tree, a bat held by a player, the dog on the right of the image, etc.). 
    Make sure that all of the captions refer to exact the same subject.
3. Exclude large background elements from your captions, such as the sky, the ground, the walls, etc.

Finally, return your final response as a JSON in the form
{
    "[subject identification]":
        [
            "[caption 1]", "[caption 2]", "[caption 3]", ...
        ],
    ...
}

You may now begin!

\end{lstlisting}

\subsubsection{Detecting Bounding Boxes}
To detect an object named \texttt{[object]} using CogVLM \cite{wang2023cogvlm}, we prompt the model with

\begin{lstlisting}
Where is [object]?
\end{lstlisting}

\section{Additional Examples of Beam Search}

\begin{figure*}[p]
    \centering
    \includegraphics[width=0.9\textwidth]{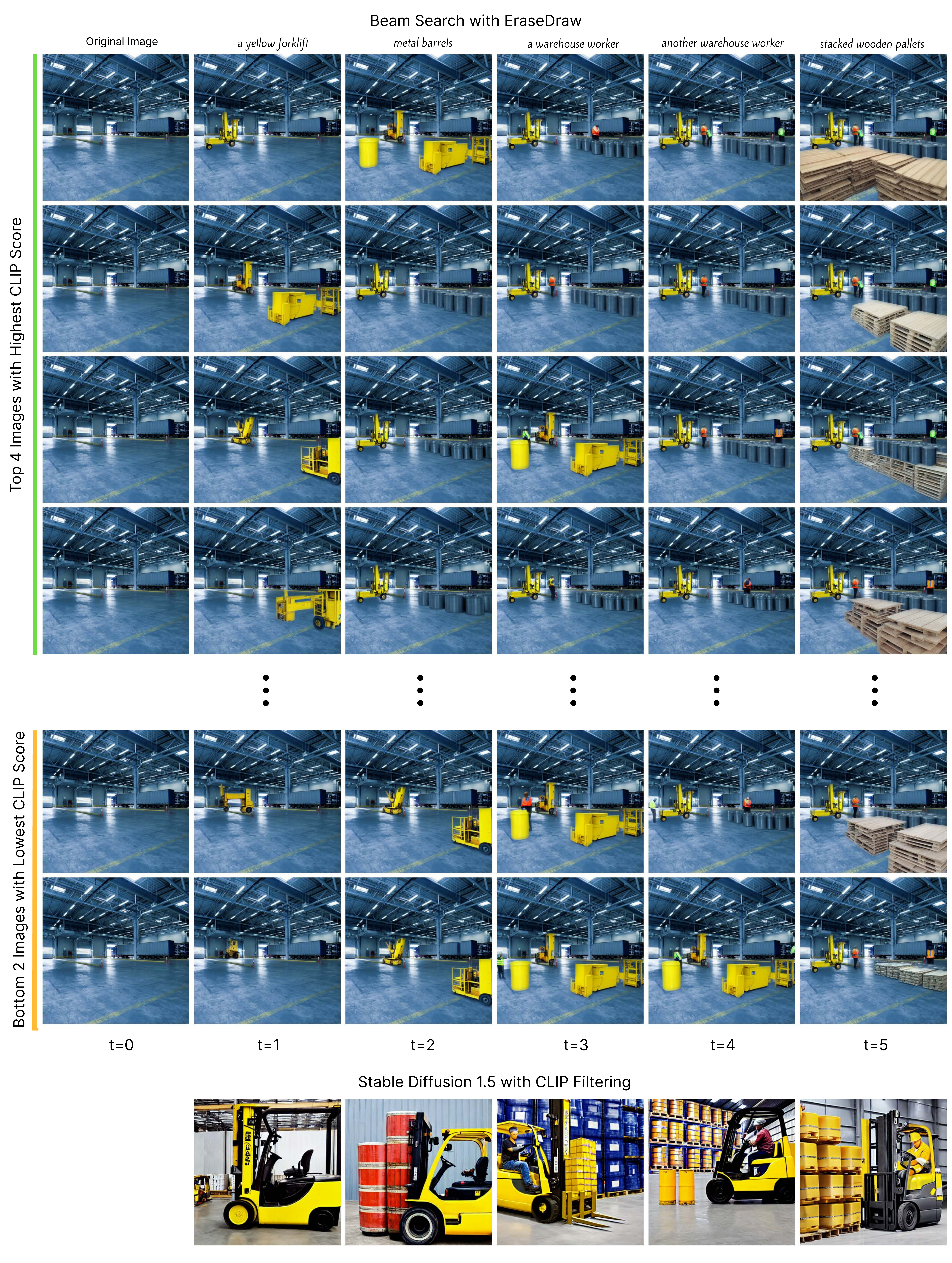} 
    \caption{Beam search with beam width $k=3$ and branching factor $N=4$ on the prompt "a yellow forklift, metal barrels, a warehouse worker, another warehouse worker, stacked wooden pallets"}
    \label{fig:iterative-generation-1}
\end{figure*}

\begin{figure*}[p]
    \centering
    \includegraphics[width=0.9\textwidth]{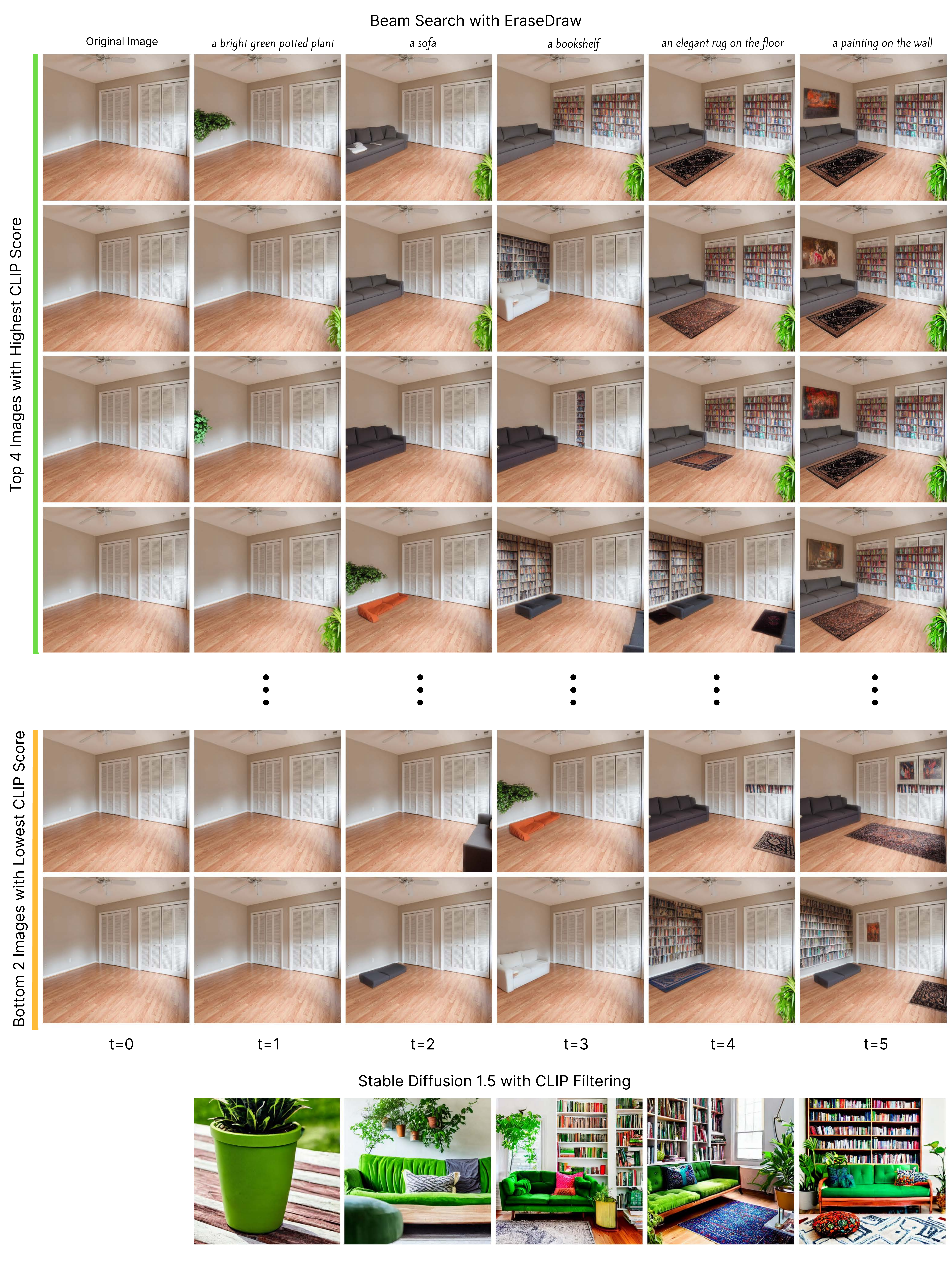} 
    \caption{Beam search with beam width $k=5$ and branching factor $N=4$ on the prompt "a bright green potted plant, a sofa, a bookshelf, an elegant rug on the floor, a painting on the wall"}
    \label{fig:iterative-generation-2}
\end{figure*}

\begin{figure*}[p]
    \centering
    \includegraphics[width=0.9\textwidth]{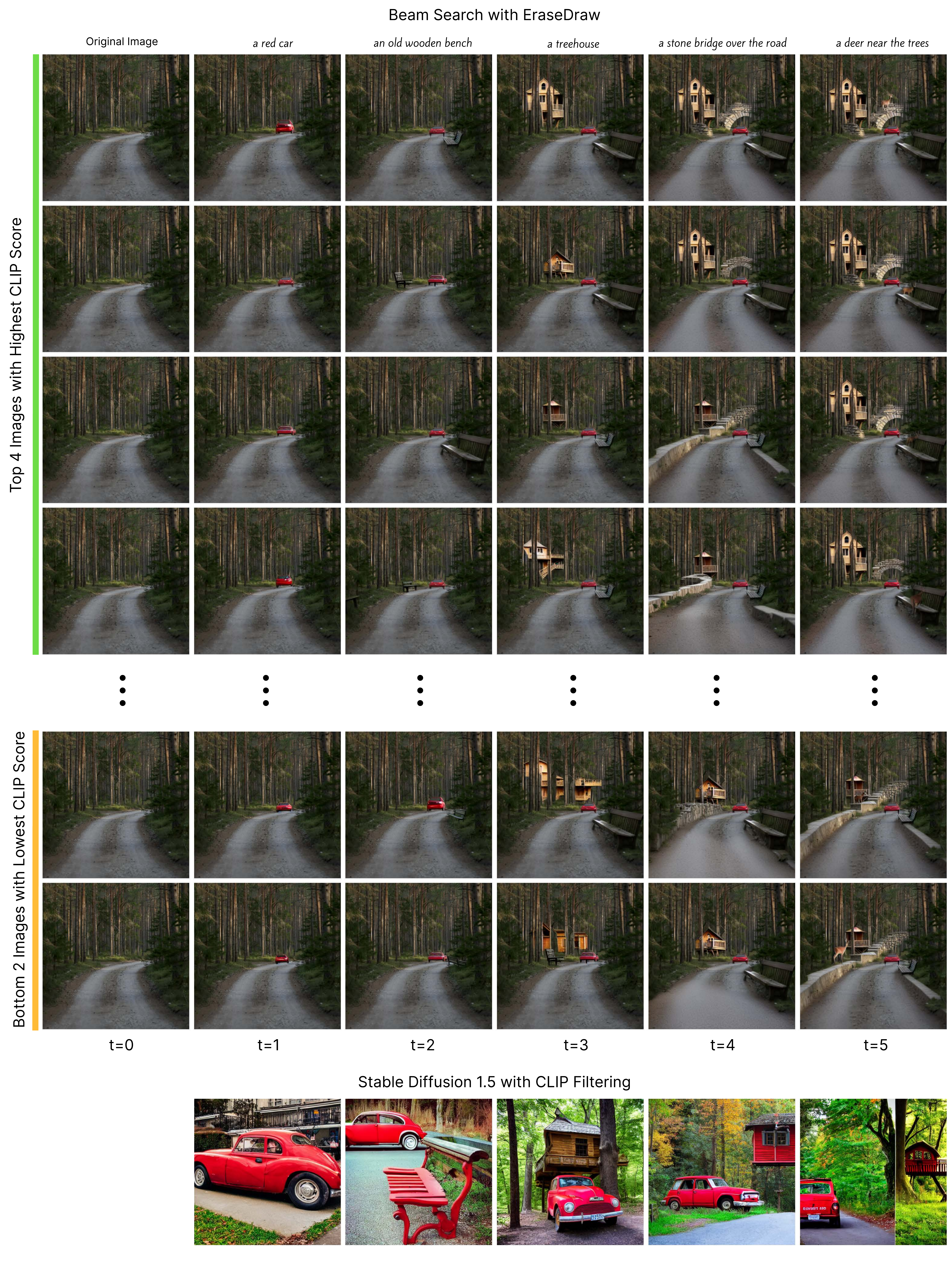} 
    \caption{Beam search with beam width $k=3$ and branching factor $N=4$ on the prompt "a red car, an old wooden bench, a tree house, a stone bridge over the road, a deer near the trees"}
    \label{fig:iterative-generation-3}
\end{figure*}

\begin{figure*}[p]
    \centering
    \includegraphics[width=0.9\textwidth]{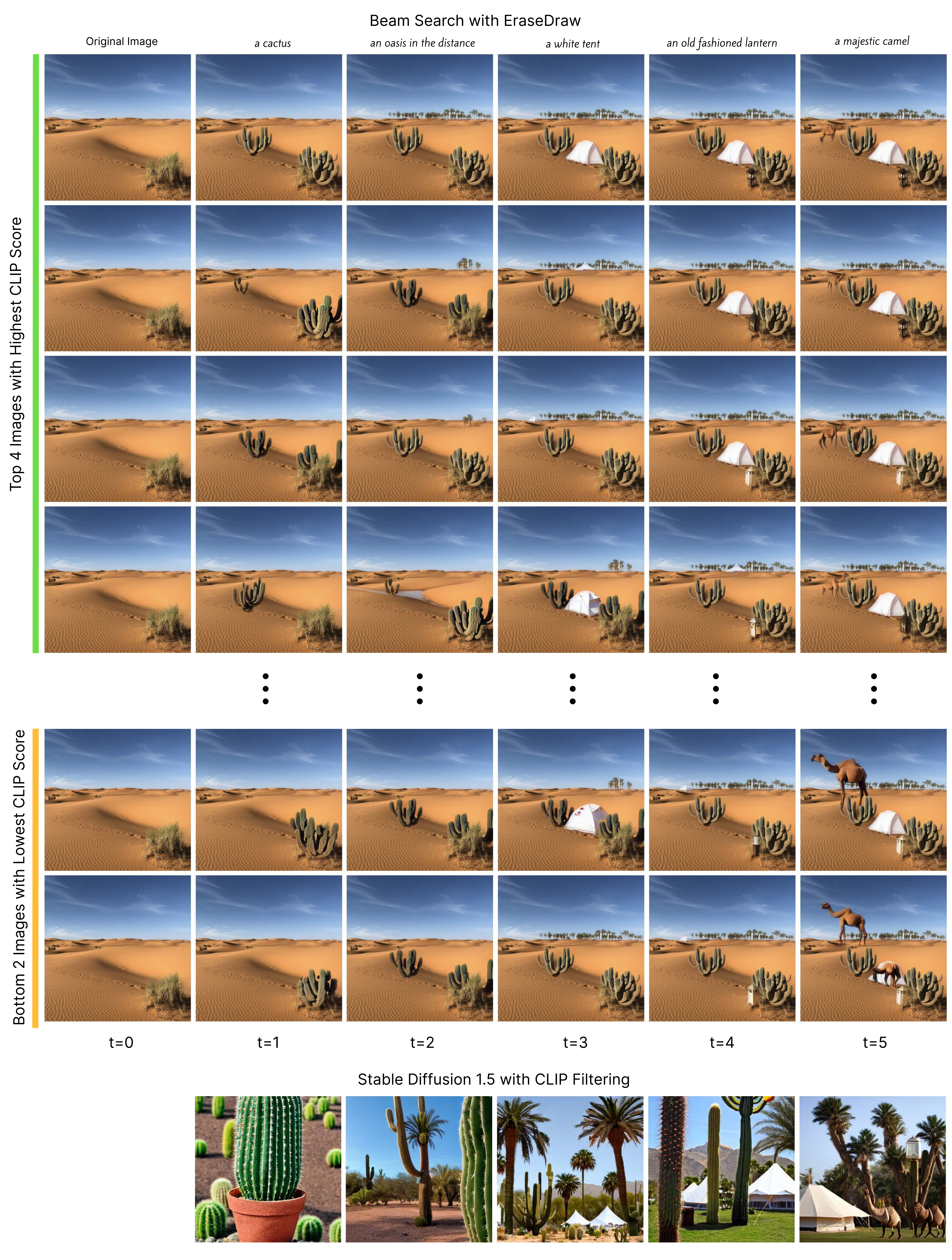} 
    \caption{Beam search with beam width $k=3$ and branching factor $N=4$ on the prompt "a cactus, an oasis in the distance, a white ten, an old fashioned lantern, a majestic camel"}
    \label{fig:iterative-generation-4}
\end{figure*}

\begin{figure*}[p]
    \centering
    \includegraphics[width=0.9\textwidth]{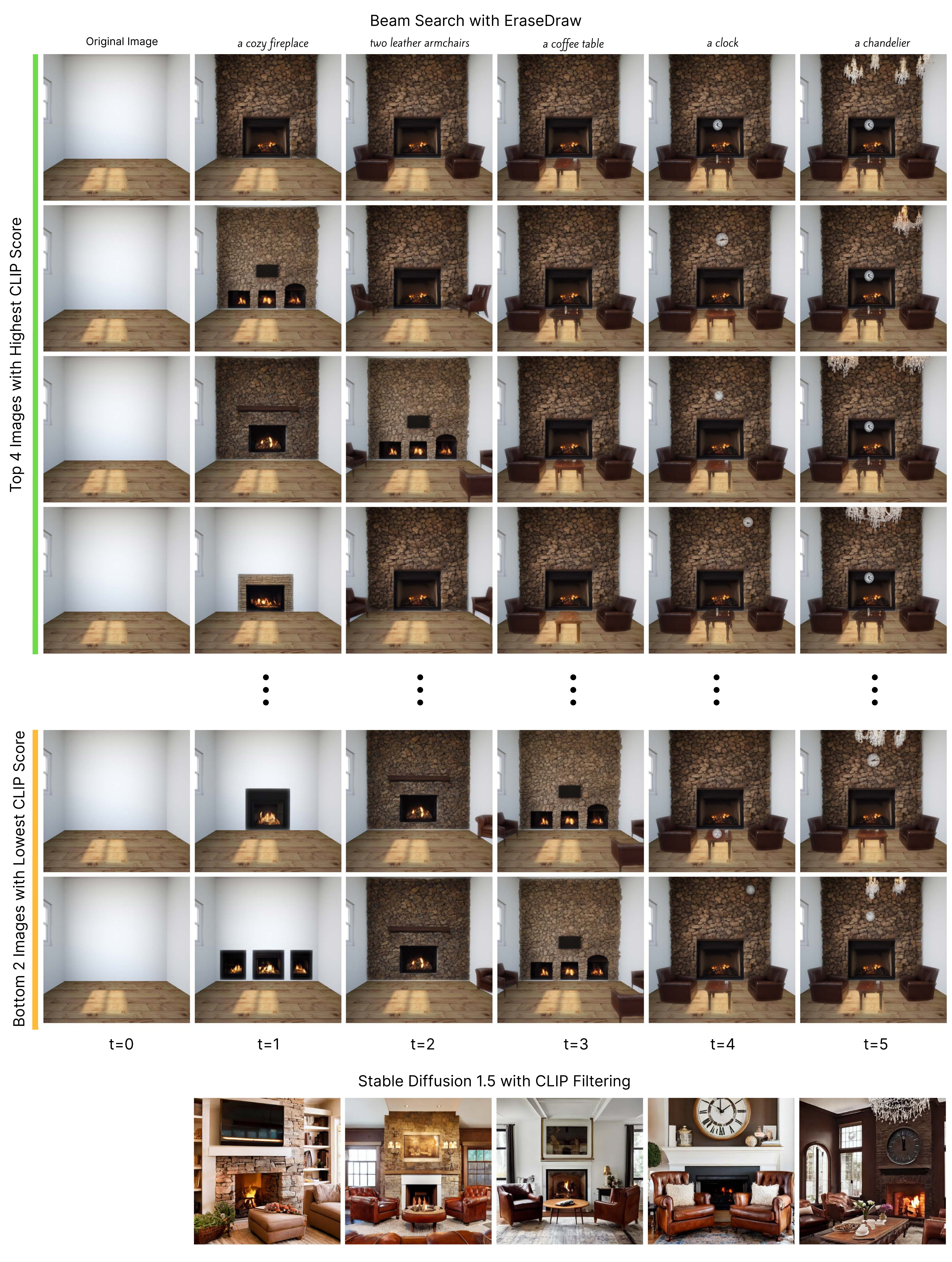} 
    \caption{Beam search with beam width $k=3$ and branching factor $N=4$ on the prompt "a cozy fireplace, two leather armchairs, a coffee table, a clock, a chandelier"}
    \label{fig:iterative-generation-5}
\end{figure*}

\begin{figure*}[p]
    \centering
    \includegraphics[width=0.9\textwidth]{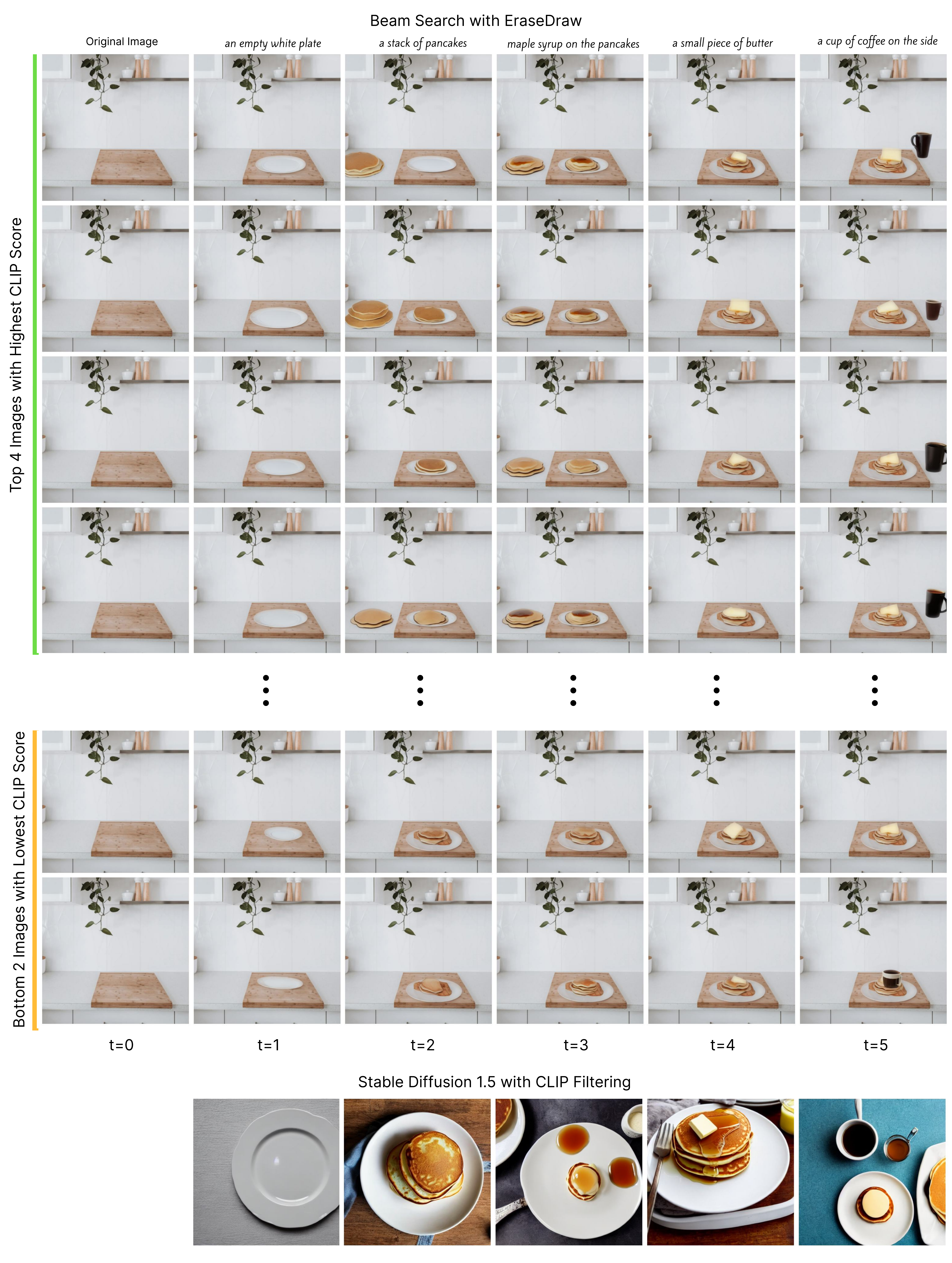} 
    \caption{Beam search with beam width $k=3$ and branching factor $N=4$ on the prompt "an empty white plate, a stack of pancakes, maple syrup on the pancakes, a small piece of butter, a cup of coffee on the side"}
    \label{fig:iterative-generation-6}
\end{figure*}

In figures \ref{fig:iterative-generation-1}, \ref{fig:iterative-generation-2}, \ref{fig:iterative-generation-3}, \ref{fig:iterative-generation-4} \ref{fig:iterative-generation-5}, \ref{fig:iterative-generation-6}, we showcase results from employing the EraseDraw method combined with beam search for gradual image composition. The complete image prompt, detailed in each figure's caption, is divided into five sub-prompts, corresponding to five steps in the beam search process. At each step \(t\), images are ranked by their CLIP similarity to the combined sub-prompts from 0 through \(t\).

Each figure's top six rows display images with the highest and lowest four CLIP scores. The last row shows a comparison with images generated by Stable Diffusion 1.5. To ensure fairness in terms of computation budget, we allow Stable Diffusion to generate as many images as beam search at each step, and apply CLIP filtering to get the best image. More specifically, for each step \(t\) in the process, Stable Diffusion 1.5 generates $(\text{beam width}) \times (\text{branching factor}) \times t$ images. These images are generated using the cumulative prompt, which is the combination of sub-prompts from 0 through \(t\). We then select the image with the highest CLIP score relative to the cumulative prompt for presentation.

We compute all CLIP scores using \texttt{siglip-base-patch16-224} \cite{zhai2023sigmoid} model from Huggingface.

\section{Dataset Generation Algorithm}

We outline our precise dataset generation procedure below in Algorithm \ref{alg:dataset}.
\begin{algorithm}[htbp]
\caption{Autonomous Data Generation}
\label{alg:dataset} 
\begin{algorithmic}[1]
\State \textbf{Input:} Input dataset $D$ of unlabeled images 
\State \textbf{Output:} Dataset $D'$ of $(C_{Image}, C_{Text}, \bm{x})$ tuples.
\State Initialize $D' \gets \emptyset$
\For{each image $\bm{x}$ in $D$}
\State Use image captioner to propose objects $\bm{o}_i$, simple captions $T_i$ for each object, and a set of complex captions $\{T^{(i)}_1 \ldots, T^{(i)}_K\}$ for each object
\State $\bm{b}_i$ = DetectBoundingBox($\bm{x}$, $T_i$)
\State $\bm{m}_i$ = BoundingBoxToSegmentation($\bm{x}$, $\bm{b}_i$)
\For{each complex caption $T^{(i)}_j$}
\State $C_{Image}$ = Erase($\bm{x}$, $\bm{m}_i$) 
\State append $(C_{Image}, C_{Text}=T^{(i)}_j, \bm{x})$ to $D'$
\EndFor
\EndFor
\end{algorithmic}
\end{algorithm}


\begin{figure*}[t]
  \centering
  \includegraphics[width=\linewidth]{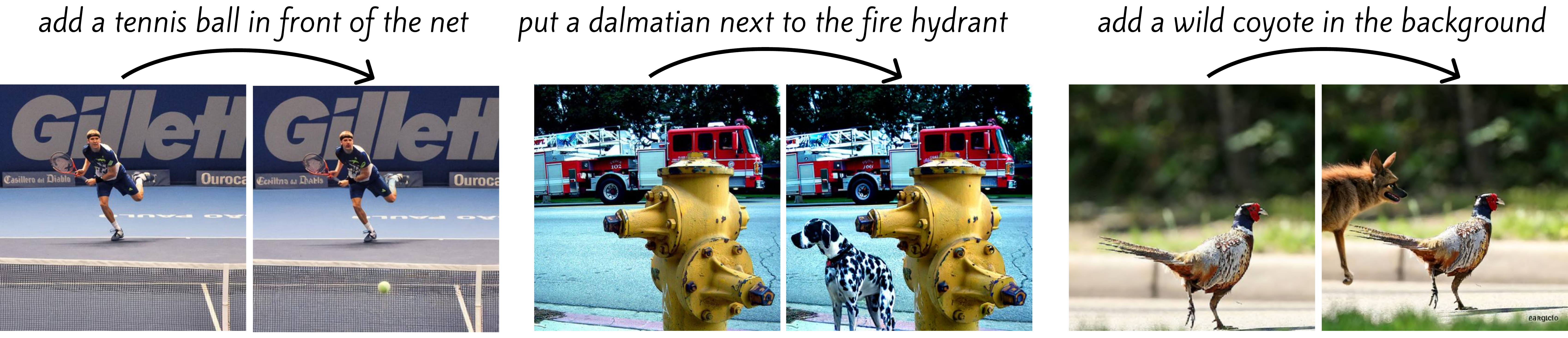}
   \caption{EraseDraw's results on insertion tasks given in Figure \ref{fig:failures} of the paper}
   \label{fig:ours-on-failure}
\end{figure*}

\end{document}